\documentclass[10pt,conference,a4paper]{IEEEtran}
\usepackage[T1]{fontenc}
\usepackage{amsfonts}
\usepackage[cmintegrals]{newtxmath}
\usepackage{url}
\usepackage{soul,color,graphicx,float,epstopdf}
\usepackage{tablefootnote,textcomp,gensymb}
\usepackage{eurosym,booktabs,multirow}
\usepackage[caption=false]{subfig}
\usepackage{amsfonts} 
\usepackage{algorithmic}
\usepackage{algorithm}
\usepackage{array}
\usepackage{stfloats}
\usepackage{float}
\usepackage{mathtools}
\usepackage{graphicx}
\usepackage{pdfpages}
\usepackage{subfig} 
\usepackage{balance}
\usepackage{diagbox}
\definecolor{Orange}{rgb}{1,0.5,0}

\makeatletter
\newcommand*{\rom}[1]{\expandafter\@slowromancap\romannumeral #1@}
\makeatother
\begin{document}
\title{Bayesian Optimisation-Assisted Neural Network Training Technique for Radio Localisation}
\author{\IEEEauthorblockN{Xingchi Liu\IEEEauthorrefmark{1},
Peizheng Li\IEEEauthorrefmark{2},
Ziming Zhu\IEEEauthorrefmark{3},
}\\ 
\IEEEauthorblockA{
\IEEEauthorrefmark{1}University of Sheffield, UK\\ 
\IEEEauthorrefmark{2}University of Bristol, UK\\
\IEEEauthorrefmark{3}Bristol Research \& Innovation Laboratory, Toshiba Europe Ltd., UK\\ 
Email: {xingchi.liu@sheffield.ac.uk}\\
{peizheng.li@bristol.ac.uk}\\
{ziming.zhu@toshiba-bril.com}}}
\maketitle
\begin{abstract}
Radio signal-based (indoor) localisation technique is important for IoT applications such as smart factory and warehouse. Through machine learning, especially neural networks methods, more accurate mapping from signal features to target positions can be achieved. However, different radio protocols, such as WiFi, Bluetooth, etc., have different features in the transmitted signals that can be exploited for localisation purposes. Also, neural networks methods often rely on carefully configured models and extensive training processes to obtain satisfactory performance in individual localisation scenarios. The above poses a major challenge in the process of determining neural network model structure, or hyperparameters, as well as the selection of training features from the available data. This paper proposes a neural network model hyperparameter tuning and training method based on Bayesian optimisation. Adaptive selection of model hyperparameters and training features can be realised with minimal need for manual model training design. With the proposed technique, the training process is optimised in a more automatic and efficient way, enhancing the applicability of neural networks in localisation.
\end{abstract}


\section{Introduction and background}
\label{sec:introduction}
Precise localisation of wireless devices is essential for many location-aware IoT applications and services, such as access control, geo-fencing, and remote monitoring. However, the performance of existing global navigation satellite systems is jeopardised in indoor scenarios. Therefore, indoor localisation has to largely rely on signal processing techniques to extract location information from the radio signal emitted or reflected from the target IoT devices. Existing radio-based localisation follows the routine of signal de-noising, feature extraction, and geometric model construction to deduce the target position. However, such schemes often fail when the property of the observed signal is impacted by the radio transmission, for example, the effect of multipath in the indoor environment \cite{kotaru2015spotfi}. 
Precise radio-based indoor localisation is still achievable with the help of dedicated wireless communication infrastructure such as the mmWave frequency-modulated continuous-wave (FMCW) radar. However, such luxury is not always available in the majority of IoT applications. Localisation techniques that can effectively work with more conventional WiFi, Bluetooth, etc., are much needed. Benefited from device cost reduction and wider promotion, Ultra-wideband (UWB) based solutions also become attractive.

Recently, facing the flourishing development of machine learning (ML), especially deep neural networks (NN), researchers on indoor localisation techniques began to embrace the approach of integrating signal processing with ML \cite{khan2018angle}\cite{li2021wireless}. Whether it is extracting critical handcrafted features from radio signals, then constructing relatively shallow NN, or directly feed the quantised radio signal into the deep NN model and train the target position estimation ability of the model in an end-to-end manner, various NN-based supervised learning methods have shown satisfactory localisation accuracy.
However, when designing a suitable NN to process radio data, the corresponding architecture or hyperparameters of the NN model must be carefully selected and adjusted to fit individual signal's characteristics, which brings inevitable extra manual workload. In addition, the generalisation ability of a trained model is a scarce property. Model re-training or hyperparameters tuning is always needed when the deployment environment changes. Although the trained NN model transfer can alleviate such burden to some extend \cite{9455237}, for hyperparameters such as the number of layers, the number of neurons in each layer, normalisation functions, activate functions and dropout ratio etc., it is usually necessary for an experienced professional to manually set them through trail and error
to obtain an optimal NN structure and satisfactory localisation results. Since training a (sometimes deep) NN with a set of hyperparameters can take long time, the process of finding suitable hyperparameters could potentially be highly time consuming, hence low efficiency.

To resolve the above challenge and develop efficient NN techniques for localisation purposes, we consider using Bayesian optimisation (BO) for automatic screening of the NN model hyperparameters. The relationship between the hyperparameters of the NN and the localisation performance (i.e., validation localisation errors) of the trained NN is modelled as a black-box function using Gauusian process (GP) regression. Then, an acquisition function is optimised to guide the hyperparameter tuning in an iterative manner along with the NN training process. BO can help to find the optimal combination of hyperparameters for the NN to produce the best localisation performance.

In this paper, we propose a novel BO-assisted NN training technique for radio localisation application. In particular, the BO process is implemented in a multi-fidelity and parallel processing manner to further accelerate the hyperparameter tuning and therefore train the NN more efficiently, where some of the hyperparameters are evaluated with less computational effort on training process than the others, while multiple hyperparameter combinations are evaluated in parallel. We also include the optimisation of the selection of input features and feature scaling methods that are used to train the NN. Numerical results based on both WiFi and UWB data sets demonstrate that the proposed scheme outperforms existing benchmark in terms of less tuning time and better localisation accuracy.

\section{Bayesian optimisation assisted NN training}
BO is a powerful tool for solving optimisation problems with unknown (black-box) functions. It learns a probabilistic mapping from inputs to outputs of the objective black-box functions as well as uses an acquisition function which determines the iterative evaluation strategy to obtain the global optimum of the objective function \cite{frazier2018tutorial}. 
\subsection{Gaussian process}
The most widely used surrogate is the GP model \cite{books/lib/RasmussenW06} which is a non-parametric Bayesian model. Define $x \in \mathbb{R}^d$ as the $d$-dimensional hyperparameters set of the NN model. The mapping between the hyperparameters (input) and the validation result (output) of the trained network based on the selected hyperparameters can be modelled as $\mathcal{GP}(\mu, k)$, where $\mu$ and $k$ represent the mean and covariance functions, respectively.

The GP is formulated given a data set of $n$ input-output pairs $D=\{X, y\}$, with $X$ being a $(n\times d)$-dimensional input vector and $y$ being a $n$-dimensional output. In the NN training problem, for any specific combination of hyperparameters $x_*$, the distribution of the training output can be fully characterised by the predicted mean and covariance of the GP, which can be calculated as 
\begin{align}
    &\mu_n(x_*)=\mu(x_*)+k_*^\intercal(K+\sigma^2I)^{-1}(y-\mu(x_*)),\\
    &\sigma^2_n(x_*)=k(x_*,x_*)-k_*^{\intercal}(K+\sigma^2I)^{-1}k_*,
\end{align}
where $K = k(X,X) $ and $k_*$ represents the covariance between $x_*$ and the training data $X$. In most cases, training a model with the same set of hyperparameters input may return different validation output. Therefore, in the prediction we include a zero-mean Gaussian noise with the variance $\sigma^2$. 

To determine GP's parameters for $\mu$ and $k$, one of the widely used approaches is the maximum likelihood estimation (MLE). The expression of the log marginal likelihood of GP can be written as
\begin{align}
    \log p(y|X,\theta,\mu)=&-\frac{1}{2}(y-\mu)^{\intercal}(K^\theta+\sigma^2I)^{-1}(y-\mu)\nonumber\\&-\frac{1}{2}\log|K^\theta+\sigma^2I|-\frac{n}{2}\log(2\pi),
\end{align}
where $\theta$ represents the parameter vector of the covariance function. The dependence on $\theta$ is made explicit by adding a superscript to the covariance matrix $K^\theta$. By maximising the log marginal likelihood, GP's parameters can be optimally learned from the existing data. As more observation data added into $D$, $\mathcal{GP}(\mu, k)$ can be updated via MLE. According to this, the black-box function's output at any $x_*$ can be better estimated, i.e., the validation result of the trained NN with any possible hyperparameter settings can be predicted.
\subsection{Acquisition functions}
Based on the knowledge from the GP, an acquisition function is optimised to determine the next evaluation point $x_\text{opt}$. The choice of acquisition function represents different strategies of choosing the next query point, such as those based on the probability of the function's output improvement as compared to the best incumbent output \cite{jones1998efficient}, or the known information of the location of the optimum according to its differential entropy \cite{hennig2012entropy}. In this work, we apply the widely used expected improvement (EI) function.
In EI, given training data set $D$ with size $n$, there exists an incumbent target $\tau_n$ which is defined as
\begin{align}
    \tau_n=\text{max}_{x\in X} f(x).
\end{align}

The aim is to find the next evaluation point which will give the highest EI as compared to the incumbent target. The EI function can be written as
\begin{align}
    \text{EI}_n(x)&\coloneqq \mathbb{E}_n[[f(x)-\tau_n]^+],\nonumber\\
    &~=\sigma_n(x)\phi(\frac{\Delta_n(x)}{\sigma_n(x)})-\Delta_n(x)\Phi(\frac{\Delta_n(x)}{\sigma_n(x)}),
\end{align}
where $\Delta_n(x)=\mathbb{E}[f(x)]-\tau_n$. Here $\phi(\cdot)$ and $\Phi(\cdot)$ denote the probability density function and cumulative density function, respectively. By maximizing the EI function, $x_{\text{opt}}$ is determined for conducting subsequent NN training and GP updates.
\begin{figure}[t]   
        \centering   
        \includegraphics[width=1\columnwidth]{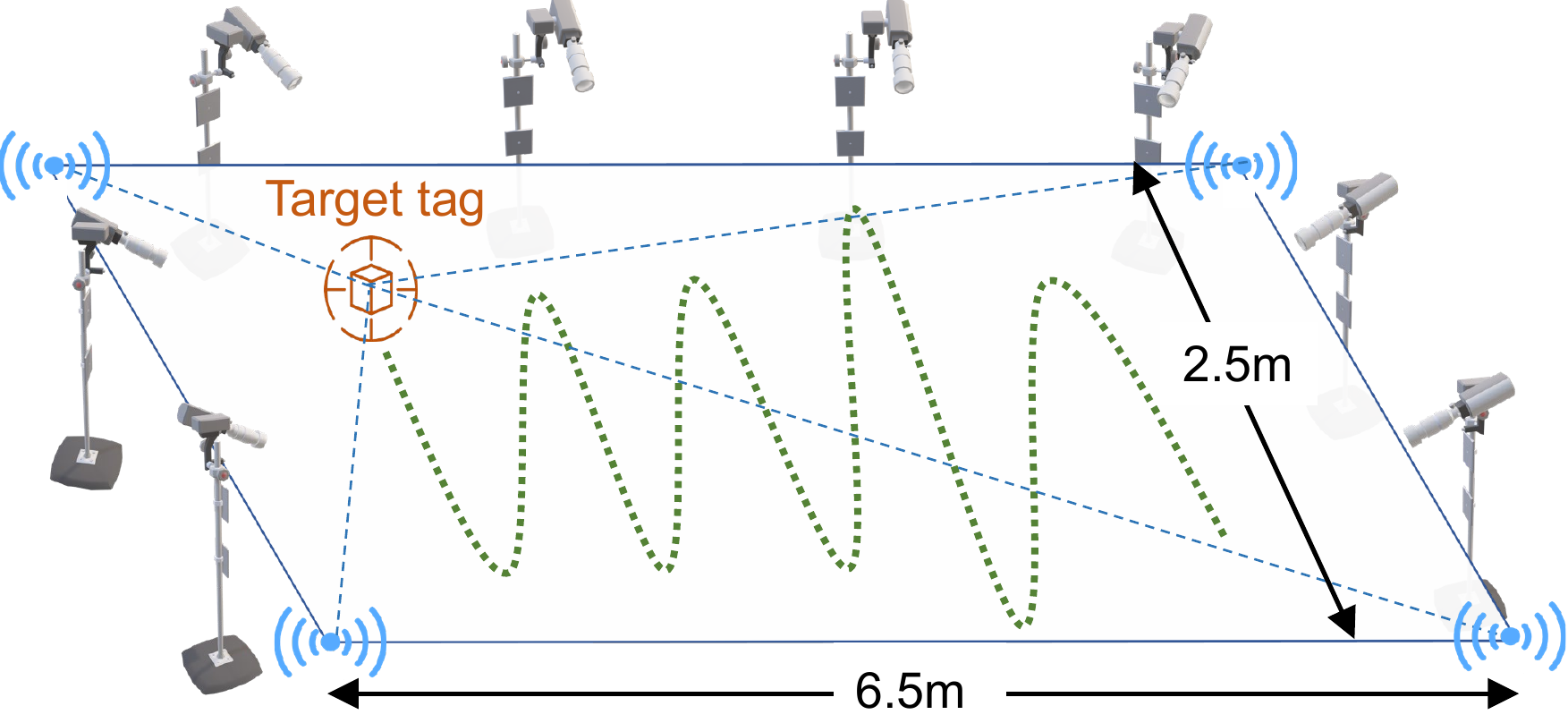}   
            \caption{The localisation scenario} \label{fig:layout}
    \end{figure} 
\section{The proposed localisation framework}
To use NN to realise radio-based localisation, the network has to be trained on a set of data containing information of signal characteristics at different locations within the environment, which is obtained through a set of experiments. For example, the experimental site used in this paper to collect data is shown in Figure \ref{fig:layout}. The experiment was conducted in an office, where a set of optical tracker system composed of 8 cameras is used to track the target to generate ground truth labels as training data. The target tag is carried by a person and moves in a curvilinear way in the venue, in which the human body is the major source of multipath effect . During this process, the wireless access points (AP) located in the corners continuously communicate with the target device and record the sounding channel characteristics as the training data. We collected data for WiFi (IEEE802.11a) and UWB systems, respectively. The WiFi data contains the channel state information (CSI) of 30 subcarriers in total. We calculated the two most possible angle of arrival (AoA) by the algorithm provided in \cite{farnham2019indoor}. The two AoAs alongside with phase and amplitude information of subcarriers contained in the CSI make up the training features of the WiFi data set. While the UWB data set features include channel impulse response (CIR), preamble symbol accumulation (PSA) and distance estimates \cite{9013500}. 
%

\begin{figure}[t]
    \centering
    \includegraphics[width=1\columnwidth]{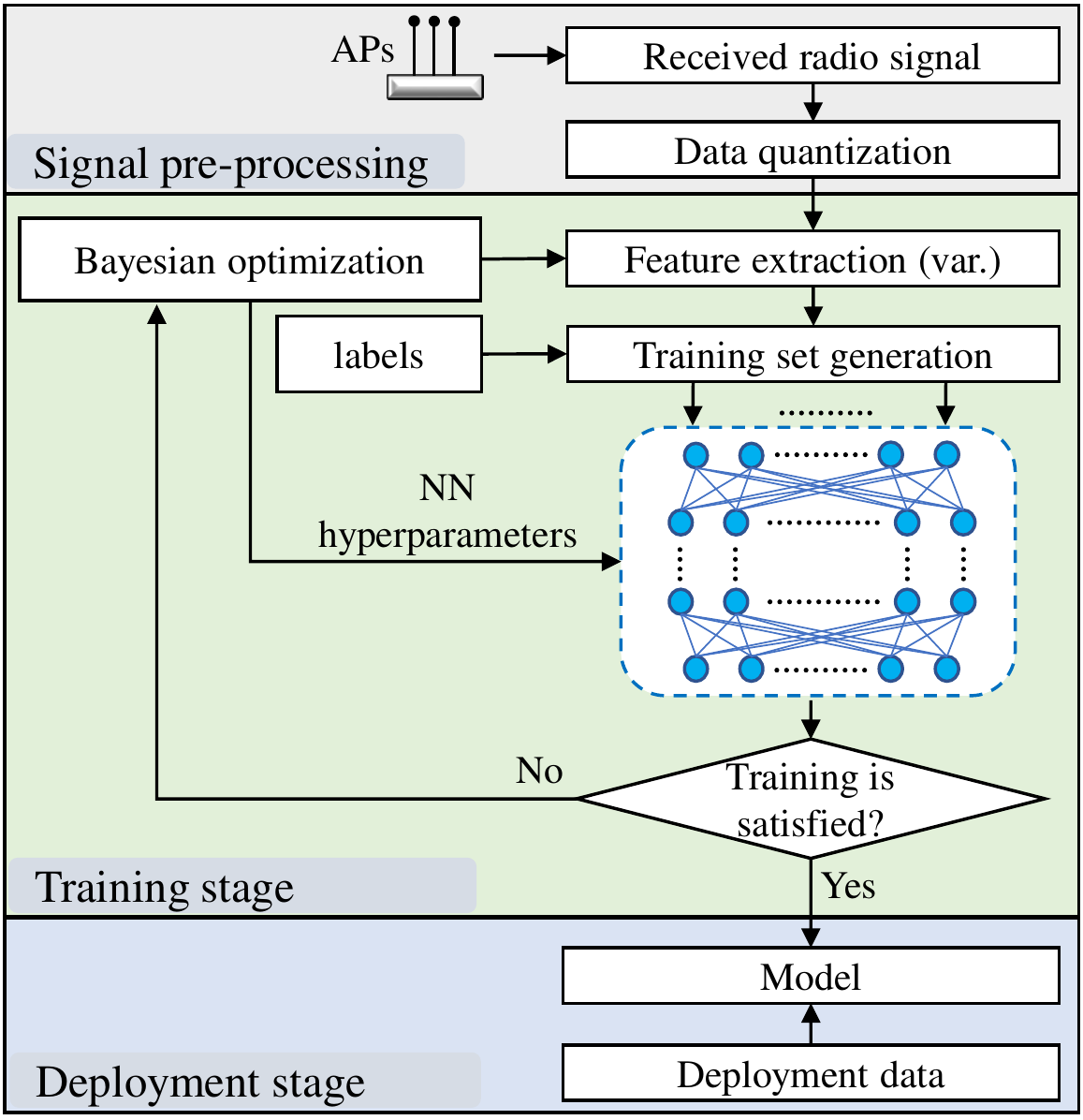} 
    \caption{The proposed BO-assisted NN model training process}
    \label{fig:flowchart}
\end{figure}

The proposed NN training process incorporating the proposed BO based hyperparameter tuning scheme is depicted in Figure \ref{fig:flowchart}. We consider the fully connection NN models in this paper. Firstly, we define a set of basic NN structure hyperparameters $x$ to be optimised, including the number of units of each layer, the dropout rate after each layer, and the learning rate. The number of extracted training features is also considered as a hyperparameter. Taking the WiFi data set as an example, we apply principal component analysis (PCA) to sort all the data features according to their eigenvalues. Then, the number of the most significant features is treated as a hyperparameter to be optimised by the BO. In addition, the choice of feature scaling method is included in $x$ as an additional hyperparameter since different methods (standardisation, normalisation, etc.) can affect the network performance. 

A special type of hyperparameter $r$ is added as the multi-fidelity factor for optimising the time and energy efficiency of the tuning process. It represents the number of training epoch required for evaluating a hyperparameter combination of $x$. With a dynamic value of $r$, some possible $x$ will be evaluated with less time and training data than the others. A median stopping rule as in \cite{46180} is used to decide how many epochs are needed for training and when to stop training. Localisation error collected by training the network 
for a small number of epochs can be treated as a low-fidelity observation of the network performance (which denotes the output of the black-box function), otherwise high-fidelity observation.
\begin{algorithm}[t]
	\caption{\color{black}Multi-fidelity BO-assisted NN training}
	\begin{algorithmic}
		\REQUIRE Prior surrogate model $\mathcal{GP}_0$, initial data $D_1$
		\WHILE {$t\leq T_{\text{budget}}$}
		\STATE Wait for one of training processes to terminate
		\STATE Observe the training epoch number $r$
		\STATE Observe the resulting validation error $f(x,r)$
		\STATE Augment data: $D_{t+1} \gets D_t \cup \left\{((x,r),f(x,r))\right\}$
		\STATE Update $\mathcal{GP}_{t+1}$ based on $D_{t+1}$
		\STATE Determine $x_{\text{opt}}$ by EI
		\STATE Start NN training with $x_{\text{opt}}$
		\ENDWHILE
		\STATE Obtain optimal hyperparameters and NN model
	\end{algorithmic}
\end{algorithm}

The tuning process could start with randomly selected hyperparameters and the corresponding training performance, as the initial data. Other criteria such as Latin hypercubes or low-discrepancy sequences can be used to generate initial data as well \cite{lee2020cost}. Then, based on the training outcome, the hyperparameters are continuously optimised by the BO throughout the tuning stage until the time budget or localisation accuracy threshold is reached. The optimised NN can then be deployed as a run-time model. The above process is described as Algorithm 1.
\section{Performance evaluation}\label{sec:Performance evaluation}
In this section, we apply the proposed scheme to tune the hyperparameters of a two-layer fully connected NN and train the network based on WiFi and UWB localisation data sets, respectively. The number of features used for training and the feature scaling method are also determined by BO. The tuning process is set to run for 200 seconds given WiFi data and 1000 seconds given UWB data. After determining the hyperparameters, the network is trained for 200 epochs. Particularly, for each case, the performance is generated by averaging the results for 10 independent simulations. The random search (RS) scheme is used as the benchmark which is widely used for tuning both numerical and categorical hyperparameters. It achieves superior performance comparing to other schemes such as grid search \cite{bergstra2012random}.

The localisation error is defined as the average Euclidean distance between the predicted location using NN and the ground truth, in metre. We define the coordinate of a point in the area as $a_i$, and $\hat{a}_i$ as the predicted location from the network. The localisation error is calculated as 
\begin{align}
    \text{Localisation error}=\frac{1}{N}\sum_{i=1}^N\sqrt{\left(a_i-\hat{a}_i\right)^2}.
\end{align}
The experiments are implemented using an Intel i5 CPU and 16 GB of memory based on AutoGluon \cite{erickson2020autogluon} and MXNet.
\vspace{-2mm}
\subsection{WiFi data set}\label{subsec:WiFi data set}
\begin{figure}[t]
    \centering
    \includegraphics[width=0.95\columnwidth]{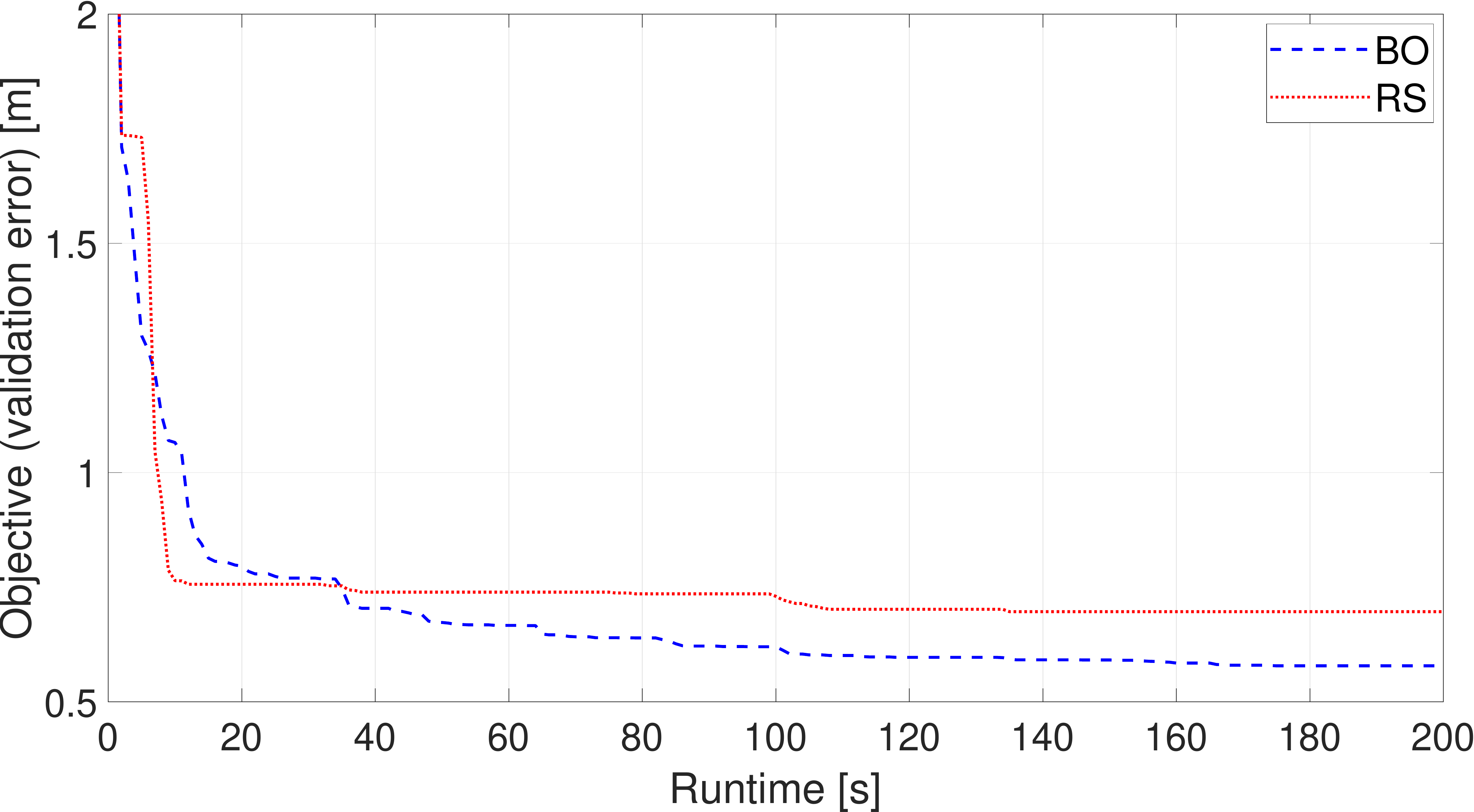}
    \caption{Tuning curve: WiFi data}
    \label{RESULT1}
\end{figure}

The search space of hyperparameters tuning is presented in Table \ref{tab 1}. In Table \ref{tab 2}, the performance of the trained NN based on the optimised hyperparameters is depicted. In Figure \ref{RESULT1}, the tuning curve shows the current best validation error versus the tuning time during the tuning process. Noted here the validation errors are collected after training the network with a only few epochs under a multi-fidelity approach. They are not the performance of a completely trained network as presented in Table \ref{tab 2}. We can see that training a NN with hyperparameters tuned by the proposed scheme outperforms the RS scheme as the resulting test error (localisation error) is reduced by 10\%. Particularly, the proposed scheme achieves the error of approximately 0.6m in less than 200 epochs, where the RS would take much longer to reach. The proposed scheme is able to achieve satisfactory performance given a relatively high-dimensional search space and can deal with both numerical and categorical hyperparameters.

\begin{table}[t]
\centering
		\caption{Hyperparameter search space}
		\label{tab 1}
\begin{tabular}{|c|c|}
\hline
\textbf{Hyperparameter}          & \textbf{Search space}                                                                                             \\ \hline
Number of units in the  $1^{\text{st}}$ layer & {[}12, 256{]}                                                                                                     \\ \hline
Number of units in the $2^{\text{nd}}$ layer & {[}12, 256{]}                                                                                                     \\ \hline
Dropout rate of the  $1^{\text{st}}$ layer    & (0, 0.75)                                                                                                         \\ \hline
Dropout rate of the $2^{\text{nd}}$ layer    & (0, 0.75)                                                                                                         \\ \hline
Learning rate                    & (1e-6, 1)                                                                                                         \\ \hline
Batch size                       & {[}16, 128{]}                                                                                                     \\ \hline
PCA number                       & {[}1, 7{]}                                                                                                        \\ \hline
Scaling method                   & \begin{tabular}[c]{@{}c@{}}L1 normalisation, L2 normalisation,\\ standardisation, minmax, no scaling\end{tabular} \\ \hline
\end{tabular}
\end{table}

\subsection{UWB data set}\label{subsec:UWB data set}
\begin{figure}[t]
    \centering
    \includegraphics[width=0.96\columnwidth]{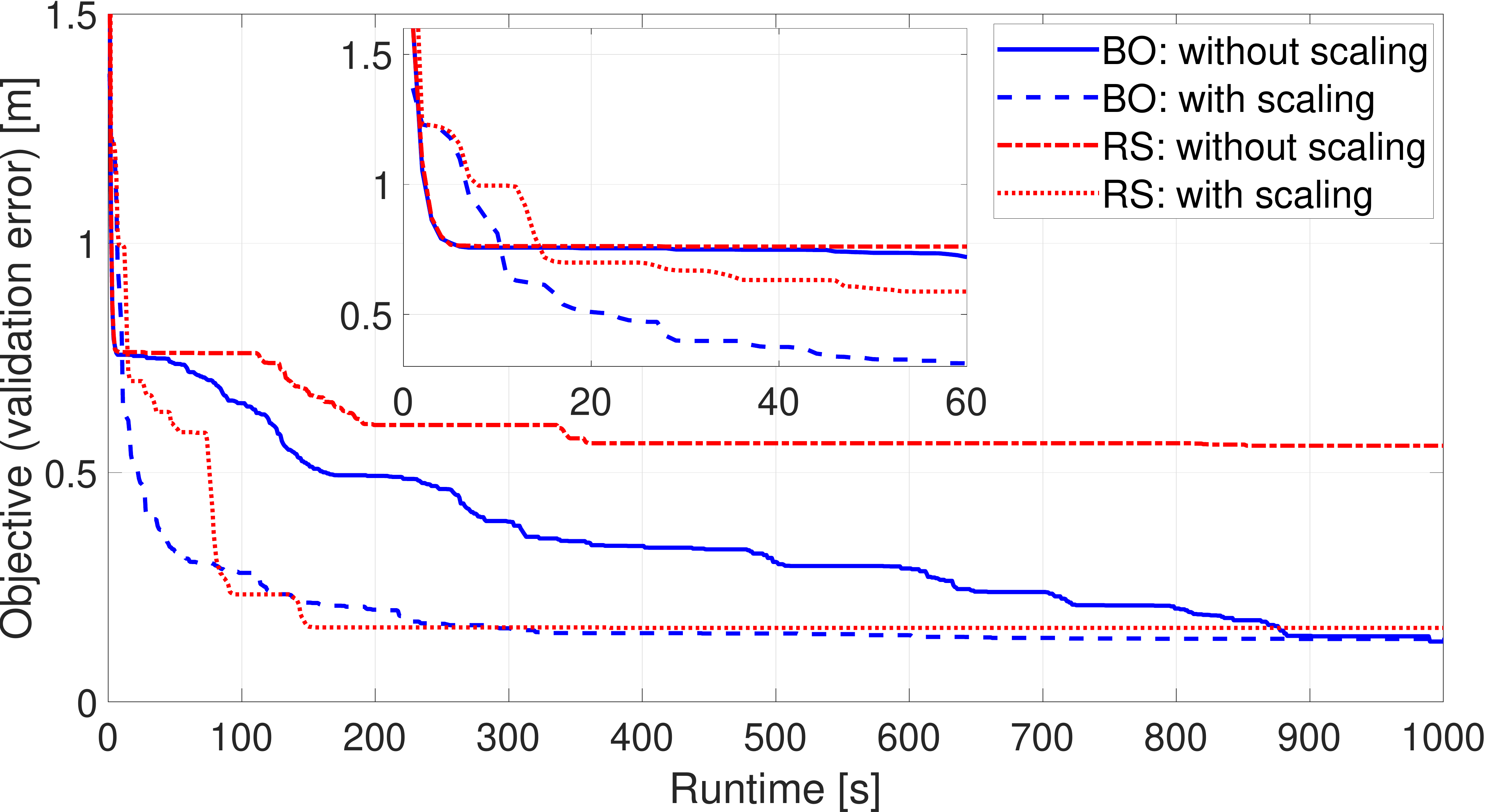}
    \caption{Tuning curve: UWB data (with enlarged result from 0 to 60 seconds.)}
    \label{RESULT2}
\end{figure}

The choice of training features of UWB data has been well screened in \cite{9013500}. We adopt the same selection in our evaluation. 
Then, the proposed scheme is utilised to jointly optimise the network hyperparameters and feature scaling method. According to Table \ref{tab 3}, using the tuning results obtained by the proposed scheme, the NN achieves 20\% lower test error as compared to the one trained based on the RS scheme. Besides, we consider training a NN using the raw data without any scaling. In this case the proposed scheme results in a 75\% test error reduction as compared to RS. The results show that the proposed scheme outperforms the RS in both settings. The tuning curves are presented in Figure \ref{RESULT2}. As seen, with and without scaling, the proposed scheme not only reaches a lower error but also achieves it faster. The BO-assisted network finds the optimal set of hyperparameters more efficiently.

\label{sec:Performance}


\begin{table}[ht]
	\begin{center}
		\caption{NN performance based on optimised hyperparameters: WiFi}\label{tab 2}
		\begin{tabular}{|c|c|c|c| p{2cm}|p{2cm}|p{2cm}}
			\hline
		\diagbox	& \textbf{Training [m]} & \textbf{Validation [m]} & \textbf{Test [m]} \\ \hline
		BO	& 0.286 & 0.531  & 0.543  \\ \hline
		RS	& 0.478 & 0.553  & 0.605  \\ \hline
		\end{tabular}
	\end{center}
\end{table}


\begin{table}[ht]
	\begin{center}
	\caption{NN performance based on optimised hyperparameters: UWB }\label{tab 3}
		\begin{tabular}{|c|c|c|c|c| p{2cm}|p{2cm}|p{2cm}}
			\hline
		\diagbox &\textbf{Feature scaling}	&\textbf{Training [m]} & \textbf{Validation [m]} & \textbf{Test [m]} \\ \hline
		\multirow{2}{*}{BO}&	Yes & 0.079 & 0.134  & 0.131  \\ \cline{2-5}
		& No 	& 0.156 & 0.103  & 0.108  \\ \hline
		\multirow{2}{*}{RS}&	Yes& 0.079 & 0.155  & 0.163  \\ \cline{2-5}
		&	No& 0.396 & 0.398  & 0.433  \\ \hline
		\end{tabular}
		
	\end{center}
\end{table}


\section{Conclusion}
\label{sec:Conclusion}
This paper proposes a NN model hyperparameter tuning and training method based on BO with a multi-fidelity approach. The proposed scheme can automatically tune the model hyperparameters to ensure the localisation accuracy and improve training efficiency while reducing the requirement of prior knowledge of NN design, which significantly improves the overall efficiency of using ML for localisation. Moreover, this BO-assisted ML scheme can be further applied to various types of radio-based localisation, and potentially be extended to a wider range of data analytical applications.

\bibliographystyle{IEEEtran} %
\bibliography{IEEEabrv,refs} 
\end{document}